\documentclass[11pt]{article}

\usepackage[letterpaper]{geometry}
\usepackage{microtype}
\usepackage{newtxtext}
\usepackage{newtxmath}
\usepackage{graphicx}
\usepackage[colorlinks=true,allcolors=blue]{hyperref}

\title{What is mathematics now, and what should it be?}

\author{Jeremy Avigad}

\date{August 24, 2026}

\begin{document}

\maketitle

\begin{abstract}
Advances in neural theorem provers have been impressive, but the successes obscure a broader vision of what AI can do for mathematics and how mathematicians can engage with AI. This essay advances a more expansive and optimistic point of view.
\end{abstract}

\section{Worries about AI}

Over the last few months, advances in large language models' ability to prove theorems, both formally and informally, have been astounding. OpenAI has announced AI-generated solutions to longstanding open problems, and many mathematicians now routinely call on systems like ChatGPT and Claude to help them prove theorems. Mathematicians' reactions to the AI revolution vary, but for students and early-career researchers in particular, the future feels dangerously uncertain.

The popular media has stoked anxiety. In the summer of 2023, a headline on the front page of the New York Times declared that ``A.I. is coming for mathematics, too.'' A year later, in the wake of Google DeepMind's success at developing a system that could solve problems in the International Mathematical Olympiad, another Times headline advised ``Move over, mathematicians, here comes AlphaProof.'' In late June of this year, a headline in IEEE Spectrum wondered ``What it means to be a mathematician when AI does the math,'' and a day later, a headline in New Scientist proclaimed bluntly, ``A golden age of maths is dawning and mathematicians are freaking out.''

While some worry that AI will eliminate jobs for mathematicians, people I talk to seem more worried about how AI will change the phenomenology of doing mathematics. Most of us are drawn to the subject simply because we enjoy thinking. We look back wistfully on our glory days as graduate students, when we had all the time in the world to explore new ideas. As our careers advance, we are distracted by the exigencies of teaching, advising, reviewing, and serving on academic and professional committees, but what drives us are the moments when we can still find time to get lost in a problem. That happens when we are in the shower, driving to work, waiting in line for coffee, enjoying a rare afternoon cleared of meetings, or in spontaneous conversation at a blackboard with a close colleague. Those moments are precisely what AI threatens to take away: it's hard to find the motivation to struggle with a question when ChatGPT can deliver an answer in seconds. Kyu-Hwan Lee, a leading practitioner of AI for mathematics, closed a recent workshop by describing, in personal terms, the emotional challenges of adapting to the new workflows and the changes to the subject he had fallen in love with as a student. To many, the sense of foreboding is visceral.

\section{A historical perspective}

The title of this essay is a nod to Richard Dedekind, who, in 1888, published an essay titled ``Was sind und was sollen die Zahlen?'' or ``What are numbers, and what should they be?'' In that essay, Dedekind gave an axiomatic characterization of the natural numbers as a system generated freely by an element, 1, and a successor function. Dedekind showed how to construct such a system in set-theoretic terms, foreshadowing twentieth-century axiomatic foundations. Dedekind also, strikingly, showed that any two number systems satisfying the axioms are isomorphic, a property of an axiomatic system that logicians now call \emph{categoricity}. These results supported Dedekind's answer to the question posed in the title: it doesn't matter what numbers are; we can take them to be anything that satisfies the axioms.

The essay was ahead of its time. Dedekind was trying to inaugurate a new style of doing mathematics that favored axiomatic characterization and set-theoretic constructions over explicit representations and calculation, in stark contrast to approaches championed by Leopold Kronecker, Karl Weierstrass, and others in the Berlin school of mathematics. At the turn of the twentieth century, David Hilbert was a staunch proponent of the new methods, but they also met with considerable resistance, and debates were often vitriolic. As late as 1964, Carl Ludwig Siegel expressed his distaste in a letter to Louis Mordell, describing his reaction to Serge Lang's \emph{Diophantine Geometry}.\footnote{A copy of the letter can be found in Lang's retrospective response, \url{https://doi.org/10.1515/dmvm-1994-0407}.}

\begin{quote}
When I first saw this book, about a year ago, I was disgusted with the way in which my own contributions to the subject had been disfigured and made unintelligible. My feeling is very well expressed when you mention Rip van Winkle!

The whole style of the author contradicts the sense for simplicity and honesty which we admire in the works of the masters in number theory --- Lagrange, Gauss, or on a smaller scale, Hardy, Landau. Just now Lang has published another book on algebraic numbers which, in my opinion, is still worse than the former one. I see a pig broken into a beautiful garden and rooting up all flowers and trees.

Unfortunately there are many ``fellow-travellers'' who have already disgraced a large part of algebra and function theory; however, until now, number theory had not been touched. These people remind me of the impudent behaviour of the national socialists who sang: ``Wir werden weiter marschieren, bis alles in Scherben zerf\"allt!'' [We will march on, until everything falls to pieces.]

I am afraid that mathematics will perish before the end of this century if the present trend for senseless abstraction --- as I call it: theory of the empty set --- cannot be blocked up.
\end{quote}

I have written extensively about nineteenth-century number theory, the transition to modern mathematics, and the associated foundational debates. I thought I knew the story well, but I have only recently come to appreciate how \emph{personal} the issues were to those involved. We are all now in a better position to empathize with early-twentieth-century mathematicians who opposed the new methods: it's not easy seeing mathematics become something strange and unfamiliar.

\section{What is AI for mathematics?}

A curious side-effect of the AI revolution is that mathematicians everywhere are being turned into philosophers, as we are forced to come to terms with our mathematical values. In a public lecture at the recent International Congress of Mathematicians in Philadelphia,\footnote{\url{https://www.youtube.com/watch?v=M0--ZH1lOzg}} Terence Tao encouraged the mathematical community to re-examine our goals and how to pursue them in the new regime, and the next day, Emily Riehl moderated a panel discussion on AI for mathematics that explored similar themes.

Both Tao's talk and the panel discussion focused largely on how we should adapt to neural theorem provers and their corporate hosts. I am uncomfortable with the implicit suggestion that ``AI for mathematics'' amounts to asking AI to prove our theorems, or with the concession that, now that a central part of the mathematical enterprise is being supplanted by AI, there is nothing left for us to do but look for a meaningful residue. The age of AI lets anyone dabble in mathematics, and we are already struggling to convince amateurs that our experiments with ChatGPT are better than theirs. Seeing mathematics beholden to big tech companies, expensive AI subscriptions, and server farms is unappealing, as is the thought of twiddling our thumbs while AI agents do our thinking for us.

The problem is that successes in neural theorem proving are crowding out a wider range of promising activities and technologies. I take the phrase ``AI for mathematics'' to encompass the formalization and digitization of mathematics, including the development of proof assistants, carefully curated libraries, automated reasoning tools, and new means of interaction and collaboration. I also take it to encompass symbolic AI, including the use of SAT solvers, SMT solvers, first- and higher-order provers, constraint solvers, and symbolic optimization techniques. Even restricting to machine learning, there is a lot more to AI than large language models: reinforcement learning and neural networks provide means of detecting patterns in mathematical data, discovering new phenomena, finding mathematical objects of interest, computing solutions to partial differential equations, and homing in on the parameters where interesting behavior occurs. These technologies offer dramatic opportunities for mathematical exploration and discovery, but the only thing mathematicians seem to be talking about is big tech's current sprint to get LLMs to settle as many high-profile conjectures as possible. It's sucking the air out of the room.

To be fair, other applications of AI in mathematics have had nowhere near the successes we have seen with neural provers. In the rush to impress investors, big tech companies and startups have poured billions of dollars into getting LLMs to prove theorems, orders of magnitude more than has been spent on more creative uses of the technologies. My enthusiasm for these other applications stems from the fact that they offer fundamentally new ways to synthesize data, discover patterns, search for mathematical objects, and find order and structure in complex phenomena. There is no telling how mathematics might advance if we spent just a small fraction of our time and energy on these other applications of AI to mathematics. We have a lot to learn about the technology and what it can do. We can't rely on anyone else to figure it out: computer scientists and big tech companies lack the mathematical expertise, skill, and disposition, and they certainly don't understand or appreciate mathematics the way we do. If we don't explore these new mathematical avenues, nobody will.

\section{Mathematical values}

We should keep in mind that AI is nothing more than technology, designed to serve our purposes. It is misguided to think of mathematicians as competing with AI; when we drive a car, we aren't competing to see who can go faster, and when we use a phone, we aren't competing to see who can speak louder.\footnote{Kim Morrison pointed me to a remark by Edsger Dijkstra that is apropos: ``The
question of whether a computer can think is no more interesting than the question of whether a submarine can swim.''} When it comes to our interactions with AI, the only relevant questions are how we want to use it, and how we \emph{should} use it to do what we want to do.

The social, cultural, political, and economic consequences of AI do, however, have bearing on what we take to be important. Our present ``conceptual'' view of mathematics gained ground in the early twentieth century in large part because it became clear that abstract, axiomatically characterized, infinitary structures offered powerful means of solving classical problems in number theory, analysis, and geometry. Even post-Grothendieck, it's not hard to trace a path from the most abstract questions to problems involving counting, measuring, predicting, building things, and laying out objects in space. Mathematicians may declare that the abstract questions are interesting in their own right, but that's only a half-truth. An important part of the appeal of mathematics is that it gives us not just a general understanding but also the thinking and problem-solving skills that bear on science, engineering, and other pursuits closer to our experience in the world.

It's clear that AI is going to transform science, business, government, and industry. If we get to a point where practitioners in these fields no longer rely on mathematics to understand what they are doing, the subject will lose a substantial part of its legitimacy and appeal. AI will be used to solve all sorts of problems, and if all that mathematics has to offer is the advice to rely on ChatGPT, its importance will diminish. This is the real sense in which we are in danger of being replaced: disciplines that currently look to mathematics for guidance will turn instead to AI.

Mathematics still has a lot to offer. Since ancient times, mathematicians have developed beautifully creative ways to solve hard problems and carry out complex chains of reasoning with limited computational resources. The era of AI is just another stage in our scientific and technological journey. LLMs are not the most efficient way to mechanize mathematical reasoning; imagine teaching an LLM to multiply big numbers by talking through the process, ``put down the seven, carry the three, \ldots'' Mathematical conceptualization and symbolization have always been central to developing powerful and efficient methods of reasoning, and that will not change.

Perhaps even more important are the values and attitudes that mathematicians embody. We know how to ask questions, develop abstractions, build theory, explore concepts, and push the boundaries of human understanding. We are disciplined, and we have a culture of engaging with problems deeply and not letting them go. We are not generally driven by short-term profit motives, but, rather, the desire to contribute something meaningful to human knowledge. These qualities are needed now more than ever; it is important for us to be leaders in this era of technological change, showing the world what AI can do well and how it can do it.

\section{The challenges}

Few mathematicians today have expertise in AI or formal methods beyond the ability to prompt ChatGPT. Yet a growing number of practitioners, most of them young, have invested time and effort in learning to use the new technologies. They have learned to develop formal libraries and APIs, automate common reasoning patterns, tune neural networks, run machine-learning experiments on a cluster, design reinforcement-learning systems, experiment with SAT encodings, and interpret statistics from a solver's run. They have learned to publish mathematical results in computer science venues when no other venues are available to them, and they have learned to submit to the standards and conventions of computer scientists who have little understanding of the mathematical enterprise or the interests of mathematicians. They have put up with these challenges because they care about mathematics and because, in their bones, they know that the new methods have something to offer.

Our discipline has not been kind to them. For most, engaging with new technologies has been tantamount to abandoning mathematics, since publications in computer science count for nothing on job applications. In the eyes of the mathematical establishment, whatever they are doing, it is not mathematics. Aspiring young mathematicians have therefore left PhD programs and postdocs to work for big tech and AI startups, and undergraduate math majors have learned to pivot to computer science to pursue the research they are interested in.

I have often called on colleagues to broaden their horizons and recognize the value of non-conventional contributions to mathematics. It's hard; we are steeped in our methodological standards. I have tried to answer questions like ``What is the nature of research in AI for mathematics?'' and ``Since the tools are getting easier to use, why can't we just use them as they become available?'' Conservatism is one of mathematics' superpowers; while computer scientists chase one shiny new thing after another, we rely on our values and judgments to sustain us for the long haul. However, the flexibility to adapt to social, political, cultural, and technological changes is also important, and mathematics has consistently evolved while preserving its core values.

Mathematics is changing. We need to anticipate the effects of AI and ensure that mathematics is equipped to address them. We owe it to the next generation to leave mathematics as strong as we found it, and to give our students the tools they will need to flourish. Those who have left the subject to work in AI and formal methods are fine: they are doing interesting work, earning good money, and settling into stable, fruitful careers. Many, however, are sad to have left mathematics, and miss its culture and values. The irony is that as much as we may not want to admit it, we need them more than they need us.

\section{The future}

To be clear, I am not claiming that, going forward, every mathematician has to become an expert in machine learning and formal methods. I do, however, think it will be important for future mathematicians to have basic competence with these tools. In contemporary mathematics PhD programs, graduate students are expected to acquire competence in algebra, analysis, geometry, topology, and other areas. Some will use functional analysis extensively in their careers, and some not at all, but we generally feel it is important for them to know what tools are available and what they can do. I expect the same will hold for the new technologies: some students will use computational reasoning tools extensively in their work and push the boundaries of what they can do, while others will use off-the-shelf tools minimally or not at all. Mathematics is a big tent and benefits from a diversity of skills and methods.

If mathematics is to succeed in the new age, however, it is not enough for mathematicians to be mere consumers of AI. We need to \emph{contribute} new ways of using AI to solve theoretical and practical problems. We need to be at the forefront of the revolution, developing more scalable and efficient reasoning procedures, finding new applications for AI, and leveraging deep expertise in and understanding of mathematics to make the most of the technology.

The question of how to get there is secondary; the first step is to reach a clearer consensus on what we want mathematics to be. If we maintain a fixed conception of the subject that clings to the status quo, we will be left helplessly lamenting each technological advance. If, instead, we recognize that the new technologies offer vast opportunities for exploration and growth, we'll find plenty to do.

\bigskip

\noindent {\em Acknowledgments.} I am grateful to several colleagues for encouragement, and to Johan Commelin, Bryna Kra, Kim Morrison, and Oliver Nash for suggestions and improvements.

\end{document}